\def\year{2018}\relax
\documentclass[letterpaper, usenames, dvipsnames]{article} %DO NOT CHANGE THIS
\usepackage{aaai18}  %Required
\usepackage{times}  %Required
\usepackage{helvet}  %Required
\usepackage{courier}  %Required
\usepackage{url}  %Required
\usepackage{graphicx}  %Required
\usepackage{times}
\usepackage{xcolor}
\usepackage{soul}
\usepackage[utf8]{inputenc}
\usepackage[small]{caption}

\usepackage{amsmath}
\usepackage[T1]{fontenc}    % use 8-bit T1 fonts
\usepackage{amsfonts}       % blackboard math symbols
\usepackage{nicefrac}       % compact symbols for 1/2, etc.
\usepackage{diagbox}
\usepackage{multirow}
\usepackage{booktabs}       % To thicken table lines
\usepackage{color}
\usepackage{tikz}
\usetikzlibrary{shapes,arrows}
\usetikzlibrary{external}
\usetikzlibrary{positioning,calc}

\tikzset{%
  block/.style    = {draw, thick, rectangle, minimum height = 3em,
    minimum width = 3em},
  sum/.style      = {draw, circle, node distance = 2cm}, % Adder
  input/.style    = {coordinate}, % Input
  output/.style   = {coordinate} % Output
}

\title{Parameter Transfer Unit for Deep Neural Networks}

\author{
Yinghua ZHANG, 
Yu ZHANG, 
Qiang YANG 
\\ 
Hong Kong University of Science and Technology \\
\{yzhangdx, yuzhangcse, qyang\}@cse.ust.hk
}
\begin{document}

\maketitle

\begin{abstract}
  Parameters in deep neural networks which are trained on large-scale databases can generalize across multiple domains, which is referred as ``transferability''. Unfortunately, the transferability is usually defined as discrete states and it differs with domains and network architectures. Existing works usually heuristically apply parameter-sharing or fine-tuning, and there is no principled approach to learn a parameter transfer strategy. To address the gap, a parameter transfer unit (PTU) is proposed in this paper. The PTU learns a fine-grained nonlinear combination of activations from both the source and the target domain networks, and subsumes hand-crafted discrete transfer states. In the PTU, the transferability is controlled by two gates which are artificial neurons and can be learned from data. The PTU is a general and flexible module which can be used in both CNNs and RNNs. Experiments are conducted with various network architectures and multiple transfer domain pairs. Results demonstrate the effectiveness of the PTU as it outperforms heuristic parameter-sharing and fine-tuning in most settings. 
\end{abstract}

\section{Introduction}
Deep Neural Networks (DNNs) are able to model complex function mappings between inputs and outputs, and they produce competitive results in a wide range of areas, including speech recognition, computer vision, natural language processing, etc. Yet most successful DNNs belong to the supervised learning paradigm, and they require large-scale labeled data for training. Otherwise, they are likely to suffer from over-fitting. The data-hungry nature makes it prohibitive to use DNNs in low-resource domains where labeled data are scarce. There is a gap between the lack of training data in real-world scenarios and the data-hungry nature of DNNs. 

\begin{figure}[htpb]
\centering
\includegraphics[scale=0.5]{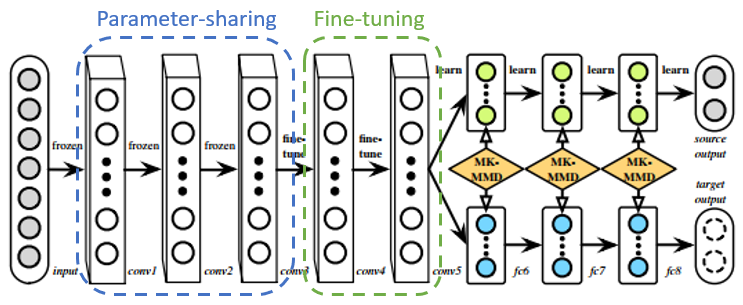}
\caption{Domain Adaptation Network (DAN) \protect\cite{long2015learning}, a typical example that applies both parameter-sharing and fine-tuning techniques. }
\label{fig:example-dan}
\end{figure}

The aforementioned dilemma can be addressed by \emph{Transfer Learning}, which boosts learning in a low-resource \emph{target} domain by leveraging one or more data-abundant \emph{source} domain(s) \cite{pan2010survey}. It is found that parameters in a DNN are \emph{transferable}, i.e., they are general and suitable for multiple domains \cite{yosinski2014transferable,DBLP:conf/emnlp/MouMYLX0J16,zoph2016transfer}. 
The generalization ability of parameters is referred as ``transferability''. Two popular parameter-based transfer learning methods are \emph{parameter-sharing} and \emph{fine-tuning}. Parameter-sharing assumes that the parameters are highly transferable, and it directly copies the parameters in the source domain network to the target domain network. The fine-tuning method assumes that the parameters in the source domain network are useful, but they need to be trained with target domain data to better adapt to the target domain. These two methods have been widely adopted. One typical example is given in Fig. \ref{fig:example-dan}, where the first three convolutional layers in a DAN are shared, and the next two layers are fine-tuned. 

Though parameter-based transfer learning by parameter-sharing and fine-tuning is prevalent and effective, it suffers from two limitations. Firstly, the parameter transferability is manually defined as discrete states, usually ``random'', ``fine-tune'' and ``frozen'' \cite{yosinski2014transferable,DBLP:conf/emnlp/MouMYLX0J16,zoph2016transfer}. But the transferability at a fine-grained scale has not been considered. A block of parameters, for example, all the filters of a convolutional layer, are treated as a whole. If they are regarded as transferable, all the parameters are retained, though some of them are irrelevant or even introduce noises to the target domain; if they are considered as not transferable, they are completely discarded and the baby is thrown out with the bathwater. 
The second limitation is that the parameter transferability differs with domains and network architectures. A parameter transfer strategy is obtained by assigning transfer states to different blocks of parameters. To find an optimal strategy, one straight-forward solution is the hold-out method. A part of the training data is reserved as a validation set. The network is decomposed into multiple parts and each part is assigned a transfer state. The optimal strategy can be found by choosing the one with the smallest validation error. Let $M$ and $L$ denote the number of transfer states and the number of parts in the network, the number of possible strategies is $M^L$. The hold-out method is rather inefficient because it involves long training time and tremendous computational costs. There is no principled approach to learn the optimal transfer strategy. 

To tackle the two limitations of existing parameter-based transfer methods, we propose a parameter transfer unit (PTU). Transfer learning with PTUs involves an already trained source domain network and a target domain network, and the two networks are connected by the PTU(s). A PTU produces a weighted sum of the activations from both networks. There are two gates in a PTU, a fine-tune gate and an update gate. The fine-tune gate adapts source domain activations to the target domain, and the update gate decides whether to transfer from the source domain. The two gates control the parameter transferability at a fine-grained scale and can be learned from data. 

The contributions of the proposed method are two folds. 
\begin{itemize}
\item A principled parameter transfer method. A novel parameter transfer unit is proposed which subsumes hand-crafted discrete transfer states and allows parameter transfer at a fine-grained scale. The unit is learned in an end-to-end approach. 
\item Plug-and-play usage. The PTU can be used in both CNNs and RNNs. It is a general and flexible transfer method as it can be easily integrated with almost all the existing models which intuitively apply the parameter-sharing or fine-tuning techniques. Experimental results show that transfer learning with the PTU outperforms the heuristic parameter transfer methods. 
\end{itemize}

\section{Related Works}
Though deep learning models have been extensively studied, there are limited research works addressing transfer learning for DNNs. The most popular transfer method for DNNs is parameter-based transfer. It is shown that parameters of the low-level layers in a CNN are transferable \cite{yosinski2014transferable}. For natural language processing tasks, Zoph \textit{et al.} and Mou \textit{et al.} study the parameter transferability in machine translation \cite{zoph2016transfer} and sentence classification tasks \cite{DBLP:conf/emnlp/MouMYLX0J16}. These works define parameter transferability as discrete states and conduct empirical studies. However, the conclusions drawn from these studies can hardly generalize to a new domain or a new network architecture. On the contrary, the proposed PTU defines transferability at a fine-grained scale and learns the transferability in a principled approach. 

Another line of research works use parameter-sharing and fine-tuning in joint with other transfer learning methods, for example, feature-based transfer learning \cite{tzeng2014deep,long2015learning,GaninUAGLLML16}. These works heuristically apply the conclusions from the empirical studies, and proposing a principled parameter-based transfer learning method is not their main focus. We expect that integrating the PTU with these models might further improve their performance. 

The most relevant work to the proposed method is the cross-stitch network \cite{misra2016cross}. Instead of assigning blocks of parameters as ``transferable'' or ``not transferable'', a soft parameter-based transfer method is adopted. Knowledge sharing between networks in two tasks is achieved with a ``cross-stitch'' unit. It learns a linear combination of activations from different networks. The proposed PTU is different from the cross-stitch network in the following two aspects. First, the transferability is controlled by linear combination coefficients in the cross-stitch unit, while the PTU learns a non-linear combination which is more expressive. Secondly, the cross-stitch unit is proposed for multi-task learning with CNNs while the PTU is designed for transfer learning where the target domain performance is the main focus. Moreover, the PTU is applied and evaluated with both CNNs and RNNs. 

\section{Parameter Transfer Unit (PTU)}
In this section, we present the proposed PTU. First, three hand-crafted discrete transfer states are introduced as background knowledge. Then we introduce the use of the proposed PTU in CNNs and RNNs. Finally, we discuss several extensions of PTU to handle the scalability issue. 

\subsection{Three Transfer States}
\begin{figure}[htbp]
\centering
\begin{tikzpicture}[node distance=2cm, 
randomnode/.style={rectangle, fill=green!15, very thick, minimum width=1.2cm, minimum height = 0.8cm},
]
%Nodes
\node[randomnode,label=below:Random] (random) {};
\node[inner sep=0pt, right of=random, fill=blue!15, minimum width=1.2cm, minimum height = 0.8cm, label=below:Fine-tune] (finetune) 
    {\includegraphics[scale=0.08]{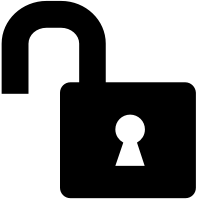}};
\node[inner sep=0pt, right of=finetune, fill=blue!15, minimum width=1.2cm, minimum height = 0.8cm, label=below:Frozen] (frozen) 
    {\includegraphics[scale=0.08]{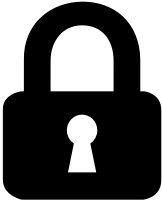}};
\end{tikzpicture}
\vspace*{-1mm}
\caption{Three hand-crafted discrete transfer states}
\label{fig:transfer-states}
\end{figure}

There are usually three states for parameter transfer, sorted in ascending order of transferability, as shown in Fig. \ref{fig:transfer-states}

\begin{enumerate}
\item Random: the parameters are randomly initialized and learned with the target domain data only; 
\item Fine-tune: the parameters are initialized with those from the source domain network, and then fine-tuned with the target domain data; 
\item Frozen: the parameters are initialized with those from the source domain network, and keep unchanged during the training process in the target domain. When parameter-sharing is applied to a convolution layer (or a RNN cell), the parameters of that layer are frozen. 
\end{enumerate}

\vspace{-0.3cm}

\vspace*{-0.5cm}
\subsection{PTU for CNNs}
\begin{figure}[htbp]
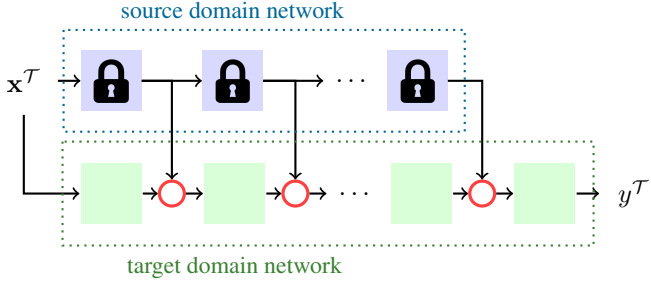

\centering
\begin{tikzpicture}[node distance=1.5cm, 
varnode/.style={circle, fill=white, very thick, minimum size=3mm},
randomnode/.style={rectangle, fill=green!15, very thick, minimum size=8mm},
gatenode/.style={circle, draw=red!75, very thick, minimum size=3mm},scale=0.9
]
\draw
	%source network
	node [varnode] (xs) {$\mathbf{x}^{\mathcal{T}}$}
	node[inner sep=0pt, fill=blue!15, minimum size=8mm, right =0.3cm of xs] (frozen1) 
    {\includegraphics[scale=0.08]{close-lock.png}}
    node[inner sep=0pt, fill=blue!15, minimum size=8mm, right = 0.8cm of frozen1] (frozen2) 
    {\includegraphics[scale=0.08]{close-lock.png}}
    node [varnode, right = 0.8cm of frozen2] (sdots) {$\ldots$}
    node[inner sep=0pt, fill=blue!15, minimum size=8mm, right = 0.1cm of sdots] (frozen3) 
    {\includegraphics[scale=0.08]{close-lock.png}}
    
    % target network
	node[randomnode, below of=frozen1] (random1) {}
    node[randomnode, right = 0.8cm of random1] (random2) {}
    node [varnode, right = 0.8cm of random2] (tdots) {$\ldots$}
    node[randomnode, right =0.1cm of tdots] (random3) {}
    node[randomnode, right =0.8cm of random3] (random4) {}
    node[varnode, right =0.3cm of random4] (yt) {$y^{\mathcal{T}}$}
    
    % transfer gates
    node[gatenode, right = 2mm of random1] (g1) {}
    node[gatenode, right = 2mm of random2] (g2) {}
    node[gatenode, right = 2mm of random3] (g3) {}
    node[MidnightBlue,above = 0.3cm of frozen2] {\footnotesize {source domain network}}
    node[OliveGreen,below = 0.3cm of random2] {\footnotesize {target domain network}}
    ;
% connections
\draw[thick,->](xs) -- node {}(frozen1);
\draw[thick,->](frozen1) -- node {}(frozen2);
\draw[thick,->](frozen2) -- node {}(sdots);
\draw[thick,->](xs) |- node {}(random1);
\draw[thick,->](frozen3) -| node {}(g3);
\draw[thick,->](random3) -- node {}(g3);
\draw[thick,->](random1) -- node {}(g1);
\draw[thick,->](frozen1) -| node {}(g1);
\draw[thick,->](g1) -- node {}(random2);
\draw[thick,->](random2) -- node {}(g2);
\draw[thick,->](frozen2) -| node {}(g2);
\draw[thick,->](g2) -- node {}(tdots);
\draw[thick,->](g3) -- node {}(random4);
\draw[thick,->](random4) -- node {}(yt);
% blocks
\draw[MidnightBlue,thick,dotted] ($(frozen1.north west)+(-0.25,0.3)$)  rectangle ($(frozen3.south east)+(0.25,-0.3)$) ;
\draw[OliveGreen,thick,dotted] ($(random1.north west)+(-0.25,0.3)$) rectangle ($(random4.south east)+(0.25,-0.3)$) ;
\end{tikzpicture}
\vspace*{-5mm}
\caption{An overview of the PTU-CNN. }
\label{fig:overview}
\end{figure}

An overview of transfer learning with PTUs in a CNN is shown in Fig. \ref{fig:overview}. The whole network, denoted by PTU-CNN, is composed of three parts, a source domain network, a target domain network and a few PTUs, which are denoted by blue blocks, green blocks and red circles in Fig. \ref{fig:overview}. A labeled target domain data sample is denoted by $(\mathbf{x}^{\mathcal{T}}, {y}^{\mathcal{T}})$. 

Let $L$ denote the number of layers in the target domain network, and the target domain network shares an identical architecture with the source domain network from the first layer to the $(L-1)$-th layer. This allows parameter transfer between different tasks or heterogeneous domains where label spaces differ. The parameters in the source domain network are frozen, and the parameters in the target domain network are randomly initialized and learned with target domain data only. PTUs are placed between the two networks in a layer-wise manner to combine activations from both domains. In the training phase, only the target domain network and the PTUs are optimized. Domain-specific knowledge can be encoded by the target domain network, and the PTUs learn how to transfer from the source domain network. In the inference phase, a target domain sample is fed into both networks, following the flows shown by the arrows in Fig. \ref{fig:overview}, and finally a predicted label is produced by the output layer of the target domain network. 

Let $l$ denote the $l$-th layer in the target domain network ($l=1, \ldots, L-1$), and $\mathbf{h}^{\mathcal{S}}_{l}$/$\mathbf{h}^{\mathcal{T}}_{l}$ denote the output of the $l$-th layer in the source/target domain network, respectively. Given $\mathbf{h}^{\mathcal{S}}_{l}$ and $\mathbf{h}^{\mathcal{T}}_{l}$, a PTU learns a nonlinear combination, denoted by $\mathbf{\tilde{h}}^{\mathcal{T}}_{l}$, and feeds $\mathbf{\tilde{h}}^{\mathcal{T}}_{l}$ to the $(l+1)$-th layer of the target domain network. 

\begin{figure}[htbp]
\centering
%\resizebox{8cm}{6cm}{
\begin{tikzpicture}[auto, thick, node distance=1.5cm, >=triangle 45, 
roundnode/.style={circle, draw=black, fill=white, very thick, minimum size=5mm}, 
varnode/.style={circle, fill=white, very thick, minimum size=5mm},
to/.style={->,>=stealth',shorten >=1pt,semithick,font=\sffamily\footnotesize},]
\draw
	% source domain classifier
	node [varnode] (xs) {$\mathbf{h}^{\mathcal{S}}_{l}$}
    node [roundnode, right of=xs] (r_gate) {$\mathbf{r}_{l}$}
    node [roundnode, right of=r_gate] (r_dot) {$\ast$}
    node[draw=black, minimum width=10mm, minimum height=8mm, above of=r_gate] (wh) 
    {$\mathbf{W}^{h}_{l}$}
    node [varnode, right of=r_dot] (xf) {$\mathbf{h}^{f}_{l}$}
    % target domain

    node [varnode, below of=xs] (xt) {$\mathbf{h}^{\mathcal{T}}_{l}$}
    node [roundnode, right of=xt] (z_gate) {$\mathbf{z}_{l}$}
%     node [block, below of=z_gate] (w) {}
    node [roundnode, below of=xf] (z_dot) {$\ast$}
%     node [varnode, below of=z_dot] (xt_tilde) {$\tilde{\mathbf{x}}^{\mathcal{T}}_{l}$}
    node [varnode, right of=z_dot] (xt_l) {$\tilde{\mathbf{h}}^{\mathcal{T}}_{l}$}
    ;
    
    \draw[->](xs) -- node {}(r_gate);
    \draw[->](xs) -- node {}(z_gate);
    \draw[->](xs) |- node {}(wh);
    \draw[->](r_gate) -- node {}(r_dot);
    \draw[->](r_dot) -- node {}(xf);
    \draw[->](wh) -| node {}(r_dot);
%     \draw[->, bend right=45] (xs)  node {}(r_dot);
	\draw[to] (xs) to[bend left=50] node[midway,below] {Identity} (r_dot);
    
	\draw[->](xt) -- node {}(r_gate);
    \draw[->](xt) -- node {}(z_gate);
    \draw[to] (xt) to[bend right=30] node[midway,below] {Identity} (z_dot);
%     \draw[->](xt) |- node {}(w);
%     \draw[->](w) -- node {}(xt_tilde);
    \draw[->](z_gate) -- node {}(z_dot);
    \draw[->](z_dot) -- node {}(xt_l);
%     \draw[->](xt_tilde) -- node {}(z_dot);
    \draw[->](xf) -- node {}(z_dot);
\end{tikzpicture}
% } %resizebox
\caption{Details of the PTU. The gates look at the activations, $\mathbf{h}^{\mathcal{S}}_{l}$ and $\mathbf{h}^{\mathcal{T}}_{l}$, from both networks. The fine-tune gate $\mathbf{r}_{l}$ decides how to adapt the source domain activations, and the update gate $\mathbf{z}_{l}$ determines how to combine the target domain activations $\mathbf{h}^{\mathcal{T}}_{l}$ with the transformed source domain activations $\mathbf{\tilde{h}}^{\mathcal{T}}_{l}$. }
\label{fig:transfer-gate-details}
\end{figure}
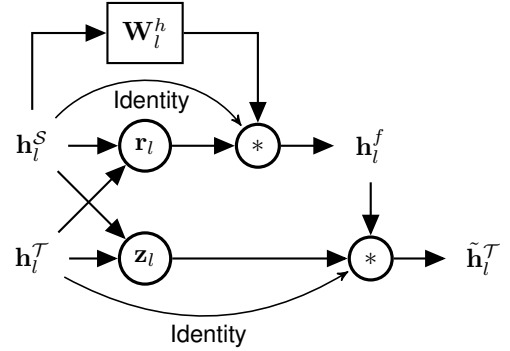

There are two gates in a PTU, a fine-tune gate $\mathbf{r}_{l}$ and an update gate $\mathbf{z}_{l}$, as defined in Eq. (\ref{eqn:gates}): 
\begin{equation}
\label{eqn:gates}
\begin{aligned}
\mathbf{r}_{l} &= \sigma (\mathbf{W}^{r}_{l} [\mathbf{h}^{\mathcal{S}}_{l}, \mathbf{h}^{\mathcal{T}}_{l}]) \\
\mathbf{z}_{l} &= \sigma (\mathbf{W}^{z}_{l} [\mathbf{h}^{\mathcal{S}}_{l}, \mathbf{h}^{\mathcal{T}}_{l}]), 
\end{aligned}
\end{equation}
where $[\cdot]$ denotes a concatenation operation and $\sigma$ denotes the sigmoid function. The gates are artificial neurons whose parameters are denoted by $\mathbf{W}^{r}_{l}$ and $\mathbf{W}^{z}_{l}$, respectively. They take the activations $\mathbf{h}^{\mathcal{S}}_{l}$ and $\mathbf{h}^{\mathcal{T}}_{l}$ as inputs, and output a value between $0$ and $1$ for each element in the activations. Then the outputs of the gates mask the hidden activations and yield the combined activation $\mathbf{\tilde{h}}^{\mathcal{T}}_{l}$, as defined in Eq. (\ref{eqn:update-with-gates}): 
\begin{equation}
\label{eqn:update-with-gates}
\begin{aligned}
\mathbf{h}^{f}_{l} &= (1-\mathbf{r}_{l}) \ast \mathbf{h}^{\mathcal{S}}_{l} + \mathbf{r}_{l} \ast \phi(\mathbf{W}^{h}_{l} \mathbf{h}^{\mathcal{S}}_{l}) \\
\mathbf{\tilde{h}}^{\mathcal{T}}_{l} &= (1-\mathbf{z}_{l}) \ast \mathbf{h}^{\mathcal{T}}_{l} + \mathbf{z}_{l} \ast \mathbf{h}^{f}_{l}
\end{aligned}
\end{equation}
where $\phi$ denotes an activation function, usually the hyperbolic tangent function or the Rectified Linear Unit (ReLU), and there is a linear transformation characterized by $\mathbf{W}^{h}_{l}$, which adapts the source domain activations to the target domain. The nonlinear transformation $\phi(\mathbf{W}^{h}_{l} \mathbf{h}^{\mathcal{S}}_{l})$ is equivalent to fine-tuning. The fine-tune gate produces a weighted sum of the source domain activations with and without fine-tuning, denoted by $\mathbf{h}^{f}_{l}$. The update gate determines how to combine the target domain activations with the transformed source domain activations. Details of the PTU are shown in Fig. \ref{fig:transfer-gate-details}. 

In extreme cases, the PTU degenerates to the hand-crafted discrete transfer states. That is, when the update gate $\mathbf{z}_{l}$ equals $0$, the fine-tune gate $\mathbf{r}_{l}$ is ignored and the activations completely come from the target domain; otherwise, it takes source domain information into consideration. When the fine-tune gate $\mathbf{r}_{l}$ equals $0$, the source domain activations $\mathbf{h}^{\mathcal{S}}_{l}$ are highly transferable and they can be directly copied to the target domain. Otherwise, transformed source domain activations are used. Thus the PTU subsumes the three discrete transfer states. In most cases, the output of the PTU is a fine-grained combination of the activations from both networks. 

\vspace*{-0.5cm}
\subsection{PTU for RNNs}
\begin{figure}[htbp]
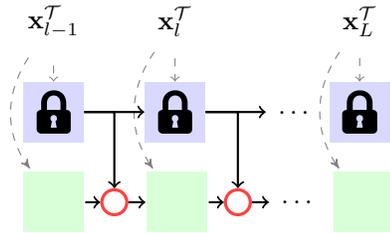

\centering
\begin{tikzpicture}[node distance=1.2cm, 
varnode/.style={circle, fill=white, very thick, minimum size=3mm},
randomnode/.style={rectangle, fill=green!15, very thick, minimum size=8mm},
gatenode/.style={circle, draw=red!75, very thick, minimum size=3mm},scale=0.9,
to/.style={draw=gray,dashed,->,>=stealth',shorten >=1pt,thin,font=\sffamily\footnotesize},
]
\draw
	%source network
	node [varnode] (xt_lm1) {$\mathbf{x}^{\mathcal{T}}_{l-1}$}
	node[inner sep=0pt, fill=blue!15, minimum size=8mm, below of = xt_lm1] (frozen1) 
    {\includegraphics[scale=0.08]{close-lock.png}}
    node[inner sep=0pt, fill=blue!15, minimum size=8mm, right = 0.8cm of frozen1] (frozen2) 
    {\includegraphics[scale=0.08]{close-lock.png}}
    node [varnode, right = 0.8cm of frozen2] (sdots) {$\ldots$}
    node[inner sep=0pt, fill=blue!15, minimum size=8mm, right = 0.1cm of sdots] (frozen3) 
    {\includegraphics[scale=0.08]{close-lock.png}}
    node [varnode, above of=frozen2] (xt_l) {$\mathbf{x}^{\mathcal{T}}_{l}$}
    node [varnode, above of=frozen3] (xt_L) {$\mathbf{x}^{\mathcal{T}}_{L}$}
    
    % target network
	node[randomnode, below of=frozen1] (random1) {}
    node[randomnode, right = 0.8cm of random1] (random2) {}
    node [varnode, right = 0.8cm of random2] (tdots) {$\ldots$}
    node[randomnode, right =0.1cm of tdots] (random3) {}
%     node[randomnode, right =0.8cm of random3] (random4) {}
%     node[varnode, right =0.3cm of random4] (yt) {$y^{\mathcal{T}}$}
    
    % transfer gates
    node[gatenode, right = 2mm of random1] (g1) {}
    node[gatenode, right = 2mm of random2] (g2) {}
    ;
% connections
\draw[draw=gray,dashed,->](xt_lm1) -- node {}(frozen1);
\draw[draw=gray,dashed,->](xt_l) -- node {}(frozen2);
\draw[draw=gray,dashed,->](xt_L) -- node {}(frozen3);
\draw[to] (xt_lm1) to[bend right=35] node[midway,below] {} (random1);
\draw[to] (xt_l) to[bend right=35] node[midway,below] {} (random2);
\draw[to] (xt_L) to[bend right=35] node[midway,below] {} (random3);
\draw[thick, ->](frozen1) -- node {}(frozen2);
\draw[thick, ->](frozen2) -- node {}(sdots);
% \draw[->](frozen3) -| node {}(g3);
% \draw[->](random3) -- node {}(g3);
\draw[thick, ->](random1) -- node {}(g1);
\draw[thick, ->](frozen1) -| node {}(g1);
\draw[thick, ->](g1) -- node {}(random2);
\draw[thick, ->](random2) -- node {}(g2);
\draw[thick, ->](frozen2) -| node {}(g2);
\draw[thick, ->](g2) -- node {}(tdots);
% \draw[->](g3) -- node {}(random4);
% \draw[->](random4) -- node {}(yt);
    
\end{tikzpicture}
\caption{Unrolled PTU-RNN. The source/target domain RNN cell is denoted by blue/green blocks. The connections from the inputs to the RNN cells are denoted by dashed gray arrows. }
\label{fig:ptu-rnn}
\end{figure}

We mainly focus on the PTU in CNNs so far, and here we extend the PTU for RNNs, denoted by PTU-RNN. As shown in Fig. \ref{fig:ptu-rnn}, a sequence $\mathbf{x}^{\mathcal{T}}$ with $L$ steps is inputted where $\mathbf{x}^{\mathcal{T}}=\{\mathbf{x}^{\mathcal{T}}_{1}, \ldots, \mathbf{x}^{\mathcal{T}}_{L}\}$. A RNN can be unrolled into a full network where the parameters in the RNN cell are shared across all time steps. Thus the RNN is able to tackle sequences of arbitrary lengths. The time step $l$ in a RNN can be regarded as the $l$-th layer in a CNN. Similarly, $\mathbf{h}^{\mathcal{S}}_{l}$/$\mathbf{h}^{\mathcal{T}}_{l}$ denotes the internal hidden state at the $l$-th time step in the source/target RNN cell, respectively. By building the relationships of the notations in CNNs and RNNs, the PTU can be readily extended to RNNs. 
% \vspace*{-0.5cm}

% \vspace*{-0.5cm}
\subsection{Scalability}
As each PTU introduces three parameters $\mathbf{W}^{r}$, $\mathbf{W}^{z}$ and $\mathbf{W}^{h}$, scalability becomes a challenge. This is because additional parameters take up more computational resources, e.g., GPU memory. In addition, more free parameters require more training data, otherwise, over-fitting is likely to occur. To reduce the computational cost, $\mathbf{W}^{r}$, $\mathbf{W}^{z}$ and $\mathbf{W}^{h}$ are shared across all time steps in the PTU-RNN. But this cannot be applied in the PTU-CNN because the dimensions of the PTU parameters in different layers do not agree. In the PTU-CNN, \emph{depth-wise separable convolutions} \cite{sifre2014rigid,howard2017mobilenets} are used instead of standard convolutions. To address the over-fitting issue, regularization techniques are necessary. Traditional regularizers, such as $l$-1 and $l$-2 regularization, together with \emph{structured sparsity} \cite{wen2016learning} are applied. These techniques allow the PTU to scale up to very deep CNNs, for example, the $28$-layer MobileNets \cite{howard2017mobilenets}. 

\subsubsection{Depth-wise Separable Convolution}
The depth-wise separable convolution is initially proposed in \cite{sifre2014rigid}, and it can greatly reduce computational costs with a slightly degraded performance \cite{howard2017mobilenets}. The depth-wise separable convolution factorizes a standard convolution into two steps, a depth-wise convolution and a $1 \times 1$ point-wise convolution. In the depth-wise convolution step, a single filter is shared across all the channels. And then the point-wise convolution applies a $1 \times 1$ convolution to the output of the depth-wise convolution. For a convolution layer with $N$ filters whose filter size is $K \times K$, the reduction in computational cost is $\frac{1}{N}+\frac{1}{K^2}$. 

\begin{figure*}[h]
\centering
\begin{minipage}{0.33\linewidth}
  \centerline{\includegraphics[scale=.18,trim={1cm 0 1cm 1.5cm},clip]{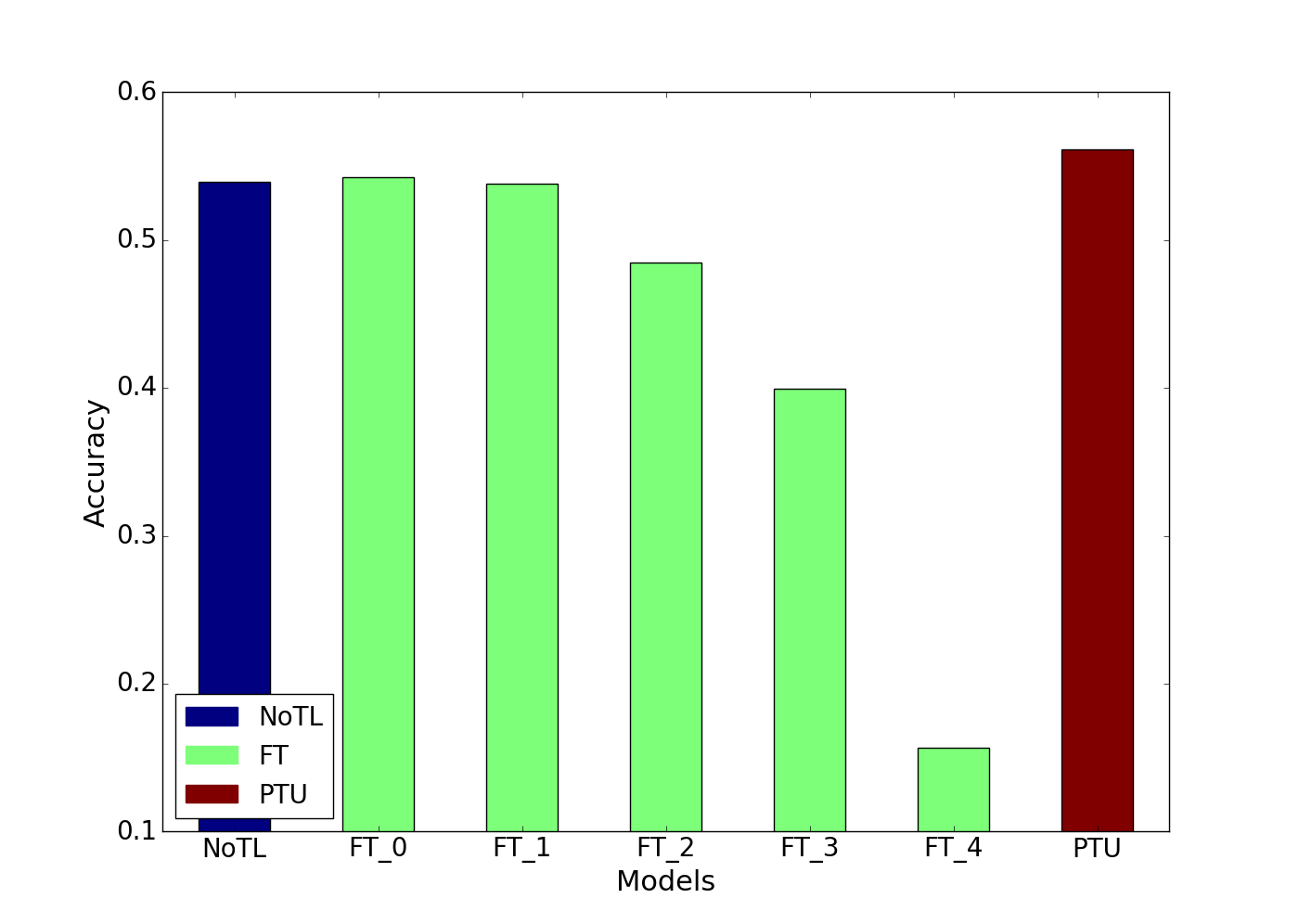}}
  \centerline{(a) S1}
\end{minipage}
\hfill
\begin{minipage}{.33\linewidth}
  \centerline{\includegraphics[scale=.18,trim={1.8cm 0 1cm 1.5cm},clip]{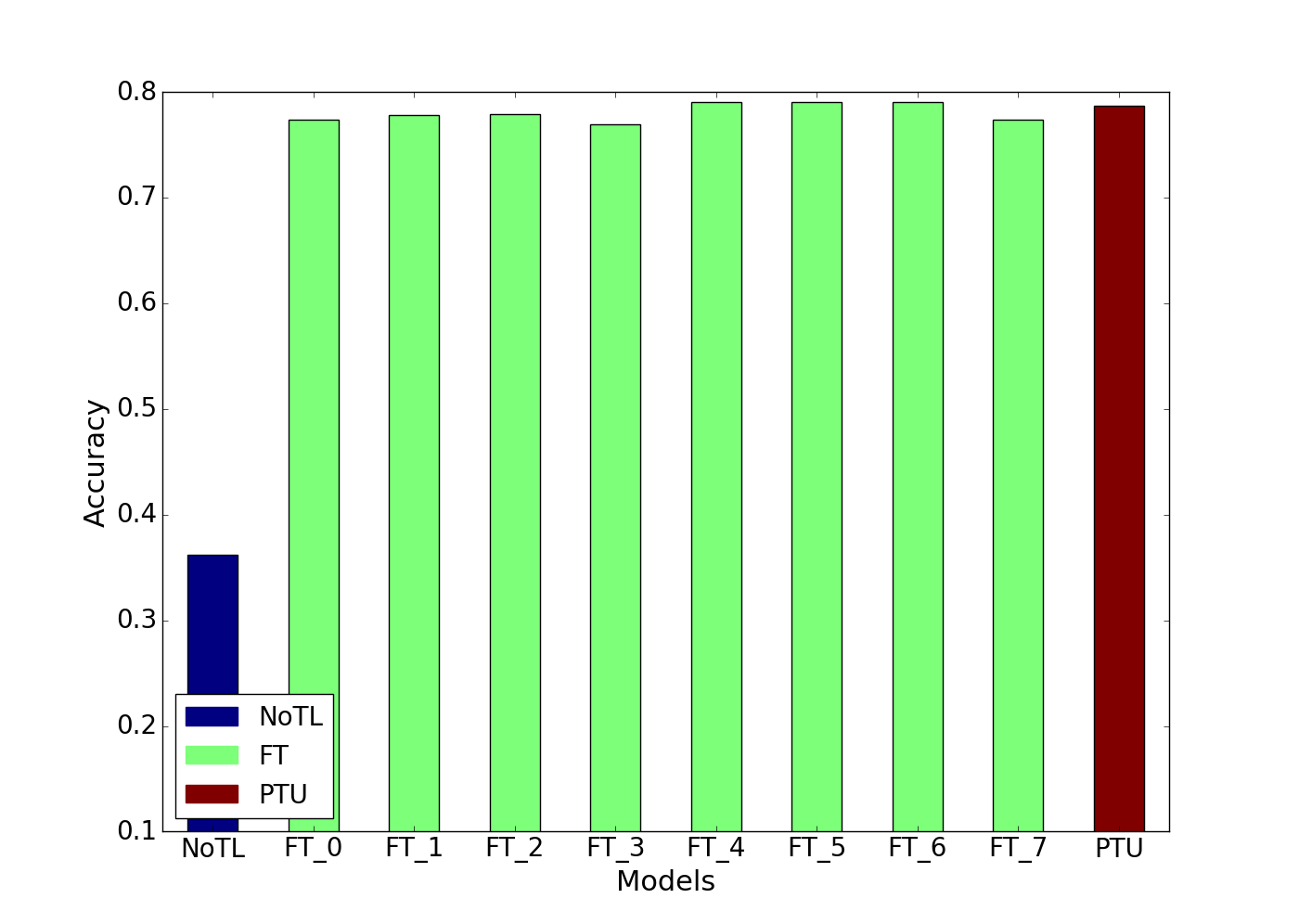}}
  \centerline{(b) S2}
\end{minipage}
\hfill
\begin{minipage}{.33\linewidth}
  \centerline{\includegraphics[scale=.18,trim={1.8cm 0.5cm 1cm 1.5cm},clip]{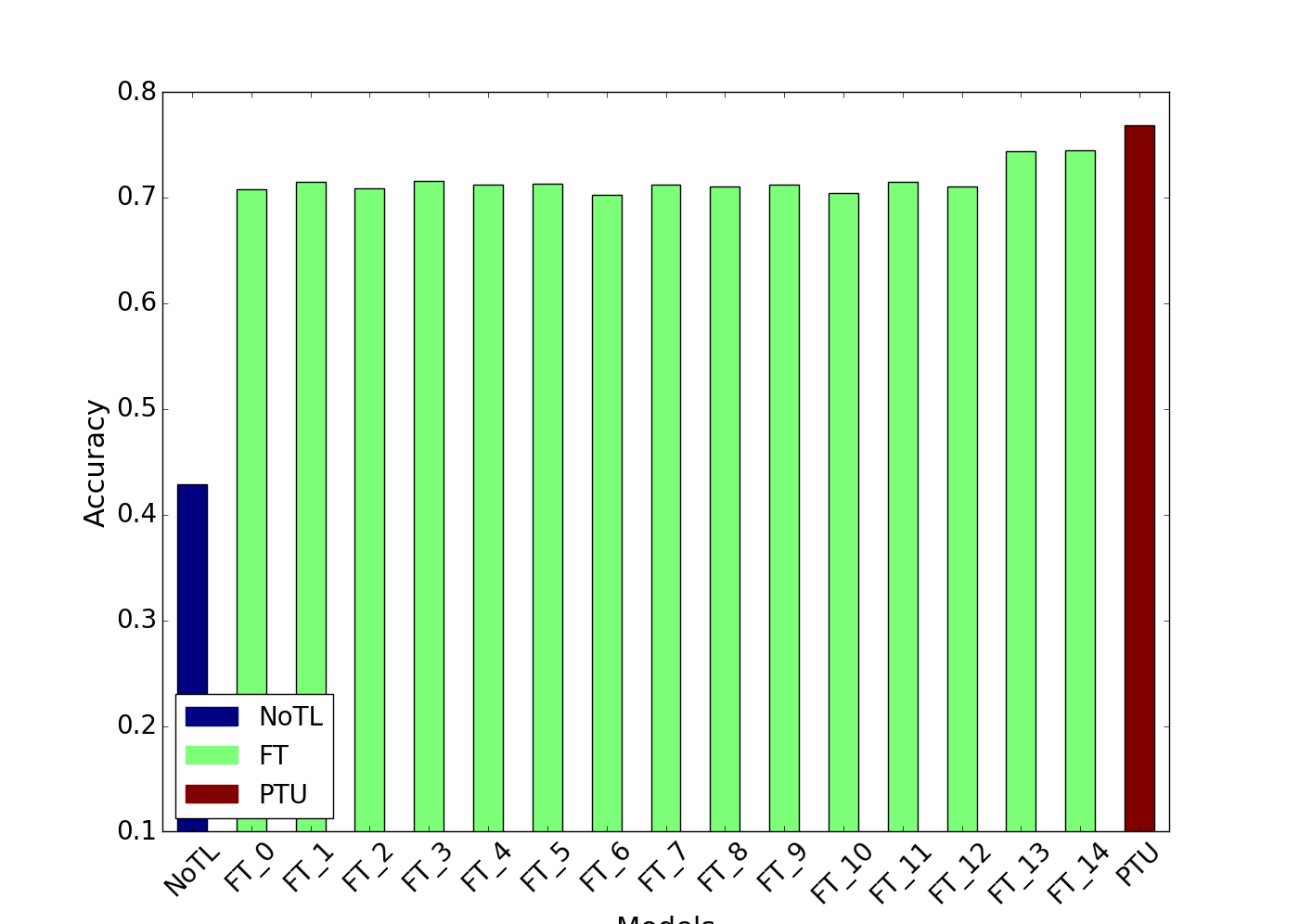}}
  \centerline{(c) S3}
\end{minipage}
\caption{Image classification accuracy of CNN models}
\label{fig:cnn-cls}
\end{figure*}

\subsubsection{Structured Sparsity}
As $\mathbf{W}^{r}$,  $\mathbf{W}^{z}$ and $\mathbf{W}^{h}$ are high-dimensional tensors, structured sparsity learning is imposed to penalizing unimportant weights and improve computation efficiency \cite{wen2016learning}. Filter-wise and channel-wise \emph{group Lasso} regularization are applied to the parameters in the PTU. 

% \vspace*{-0.2cm}
\section{Experimental Results}
We evaluate the PTU with both CNNs and RNNs on classification tasks. Classification accuracy is adopted as the evaluation metric. All the neural networks are implemented with Tensorflow \cite{abadi2016tensorflow}. 

\subsection{Experiments on CNNs}
We first describe the experimental setup, and then report numerical results. We provide an interpretation of the output values of the gates in the PTU. 
\subsubsection{Experimental Setup}
Various network architectures are evaluated on multiple transfer domain pairs. Three transfer settings are composed from four natural image classification datasets and three network architectures, as described below: 
\begin{itemize}
\item S1: CIFAR-10 $\rightarrow$ CIFAR-100 \cite{krizhevsky2009learning} where a LeNet-like $5$-layer network is employed. 
\item S2: ILSVRC-2012 \cite{ILSVRC15} $\rightarrow$ Caltech-256 \cite{griffin2007caltech} where a VGG-16 network \cite{Simonyan14c} is employed. 
\item S3: ILSVRC-2012 $\rightarrow$ Caltech-256 where a MobilenetV1 network \cite{howard2017mobilenets} is employed. 
\end{itemize}

For CIFAR-100, there is a standard train/test split, and $10\%$ training data are reserved as a validation set. The batch size is $128$. For Caltech-256, $45$ and $15$ images are used as the training and validation set for each class, respectively. The batch size is $32$. Hyper-parameters (learning rate) are selected via the hold-out method. The learning rate is chosen from $\{0.1, 0.01, 0.001, 0.0001\}$. 

% The data set statistics are listed in Table \ref{tab:dataset}. 

% \begin{table}[htbp]
% \centering
% \caption{Public datasets and their statistics}
% \label{tab:dataset}
% \begin{tabular}{|l|c|c|c|c|}
% \hline
% \diagbox[width=2.5cm]{Dataset}{Stat} & \# train  & \# valid & \# test & \# class \\ \hline
% CIFAR-100 & $1,020$ & $1,020$ & $6,149$ & $100$ \\ \hline
% Caltech-256 & & & & \\ \hline
% MIT-67 & $5,360$ & $8,920$ & $1,340$ & $67$ \\ 
% \hline
% \end{tabular}
% \end{table}

Two baseline models are considered: 
\begin{itemize}
\item No transfer (NoTL). The parameters are learned from scratch in the target domain. 
\item Layer-wise fine-tuning (FT). For a CNN with $L$ layers, if a layer has two possible transfer state, ``fine-tune'' and ``frozen'', there are $2^L$ transfer strategies, which generates prohibitive computation costs. To improve the efficiency, we adopt a strategy that layers are incrementally frozen as the parameter transferability drops when moving from low-level layers to high-level layers in a CNN \cite{yosinski2014transferable}. That is, there are $L$ fine-tuning strategies, and FT-$l$ denotes a strategy that freezes the first $l$ layers and fine-tunes the remaining layers. 
\end{itemize}

\begin{figure*}[htpb]
\centering
\begin{minipage}{0.33\linewidth}
  \centerline{\includegraphics[scale=.3,trim={0.5cm 0cm 0cm 1.2cm},clip]{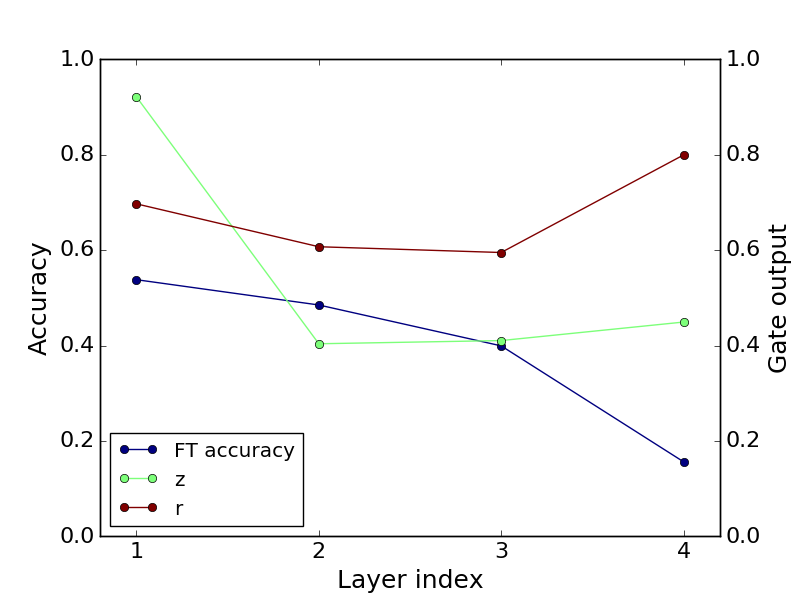}}
  \centerline{(a) S1}
\end{minipage}
\hfill
\begin{minipage}{.33\linewidth}
  \centerline{\includegraphics[scale=.3,trim={0.5cm 0cm 0cm 1.2cm},clip]{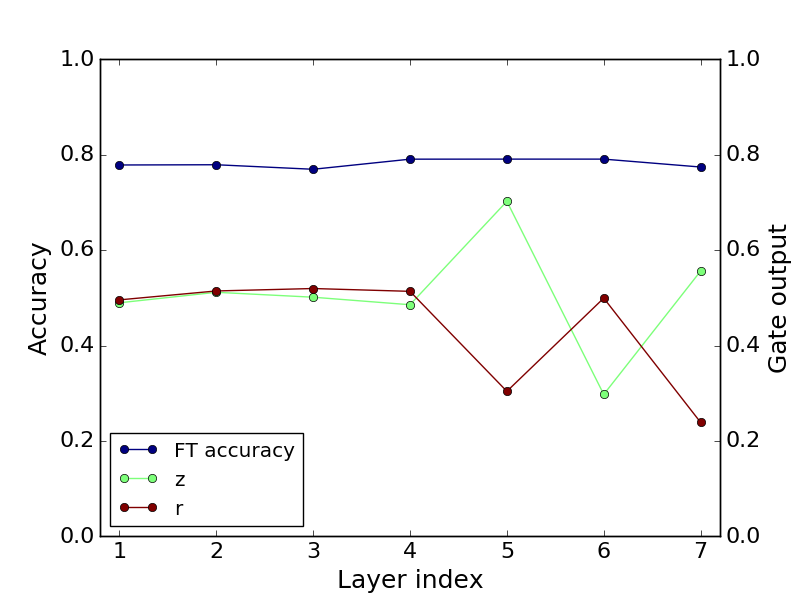}}
  \centerline{(b) S2}
\end{minipage}
\hfill
\begin{minipage}{.33\linewidth}
  \centerline{\includegraphics[scale=.3,trim={0.5cm 0 0cm 1.2cm},clip]{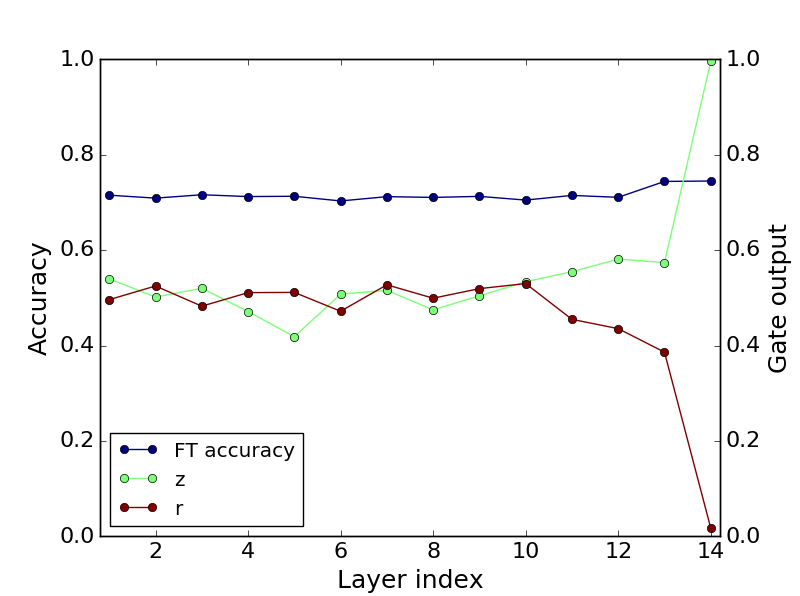}}
  \centerline{(c) S3}
\end{minipage}
\caption{Parameter transferability of different layers in CNNs}
\label{fig:cnn-gate-trans}
\end{figure*}

\subsubsection{Results}
The results of the three methods are depicted in Fig. \ref{fig:cnn-cls}. As there are $L$ fine-tuning models, only the highest test accuracy is summarized in Table \ref{tab:cnn-cls}. $\Delta$ denotes the relative improvement of the PTU model over the FT model. 

As shown in Fig. \ref{fig:cnn-cls}, the parameter transferability differs with domains and architectures. For S1, the parameters in low-level layers are more transferable than those in high-level layers, and the parameter transferability decreases monotonously, which is generally consistent with the conclusions in \cite{yosinski2014transferable}. But the conclusion does not generalize to S2 and S3. For example, in S3, freezing the first $13$ layers achieves a higher accuracy than fine-tuning the whole network. These results indicate that the heuristic layer-wise fine-tuning method might not yield the optimal parameter transfer strategy. 

Low-resource target domains benefit from parameter-based transfer learning, as both the FT model and the PTU model outperform the NoTL model. Furthermore, the PTU model achieves a comparative performance with the FT model in S2, and obtains the optimal test accuracies in the other $2$ settings with a relative improvement around $3\%$. These results demonstrate the effectiveness of the PTU in various domains and with different network architectures. Unlike the layer-wise fine-tuning which involves $L$ training processes, a reasonable parameter transfer strategy can be learned in one pass with the PTU. 

% The main focus of the proposed method is not to achieve state-of-the-art performance in a specific domain, but provide a start point with reasonable performance by parameter-based transfer learning. 

\begin{table}[htbp]
\centering
\caption{Classification accuracy on CNNs (The optimal test accuracy of a setting is highlighted with bold face.)}
\label{tab:cnn-cls}
\begin{tabular}{|c|c|c|c||c|}
\hline
\diagbox[width=2cm]{Settings}{Models} & NoTL & FT & PTU & $\Delta$(\%) \\ \hline
S1 & $0.5392$ & $0.5428$ & $\mathbf{0.5612}$ & $3.39$ \\ \hline
S2 & $0.3621$ & $\mathbf{0.7909}$ & $0.7868$ & $-0.52$ \\ \hline
S3 & $0.4290$ & $0.7450$ & $\mathbf{0.7686}$ & $3.17$ \\ \hline
\end{tabular}
\end{table}

\subsubsection{Quantify Parameter Transferability by Gate Outputs}
The two gates in the PTU control the parameter transferability, which provides an approach to quantify the parameter transferability. The average output values of the two gates in different layers are shown in Fig. \ref{fig:cnn-gate-trans}. The classification accuracy of the FT model which is an indicator of the parameter transferability is also included. 

For the update gate $\mathbf{z}$, it controls how much knowledge is flowed from the source domain to the target domain. A larger $\mathbf{z}$ indicates more knowledge transfer. For example, in Fig. \ref{fig:cnn-gate-trans}(a), the FT-$1$ achieves the optimal test accuracy, and $z_1$ is also the highest value among all the $4$ values. In addition to the update gate, the fine-tune gate $\mathbf{r}$ characterizes how many activations are copied from the source domain, and how many activations need to be transformed before applying to the target domain, which has not been considered by existing works. 

Here we quantify the parameter transferability as the average output values of gates which are scalars. A more fine-grained visualization analysis can be performed. For example, a large $z$ value of a filter might help us identify important patterns that are shared between domains. This might demystify DNNs which are considered as a black-box process. Since understanding the neural network via visualization analysis is not the main focus of this paper, it will be left as future works.

\begin{figure*}[htbp]
\centering
\begin{minipage}{0.24\linewidth}
  \centerline{\includegraphics[scale=.3,trim={0cm 0cm 0cm 1.2cm},clip]{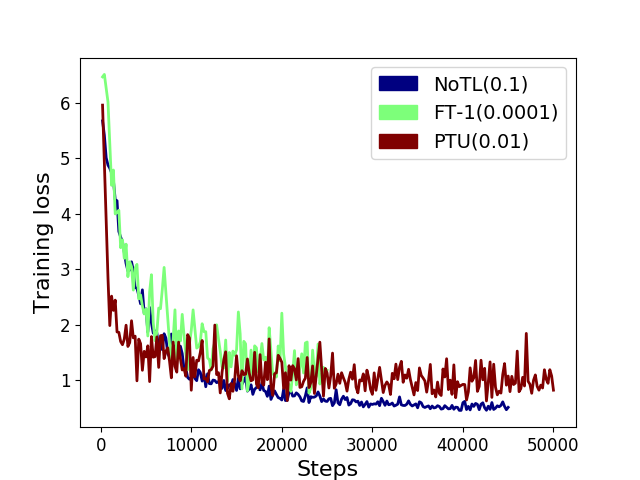}}
  \centerline{\small{(a) Training loss of S3}}
\end{minipage}
\hfill
\begin{minipage}{.24\linewidth}
  \centerline{\includegraphics[scale=.3,trim={0cm 0cm 0cm 1.2cm},clip]{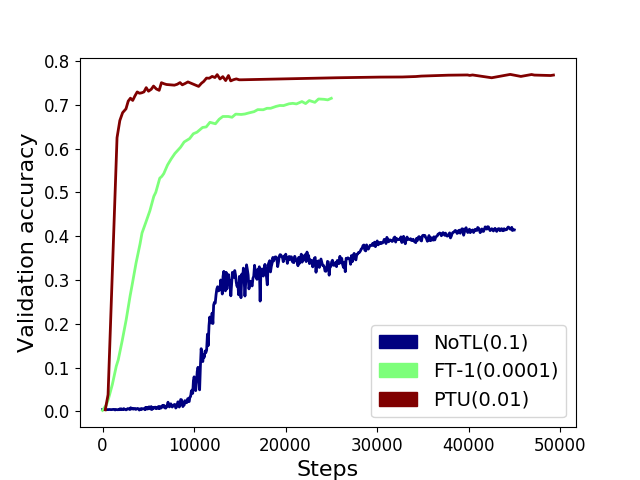}}
  \centerline{\small{(b) Validation accuracy of S3}}
\end{minipage}
\hfill
\begin{minipage}{.24\linewidth}
  \centerline{\includegraphics[scale=.3,trim={0cm 0cm 0cm 1.2cm},clip]{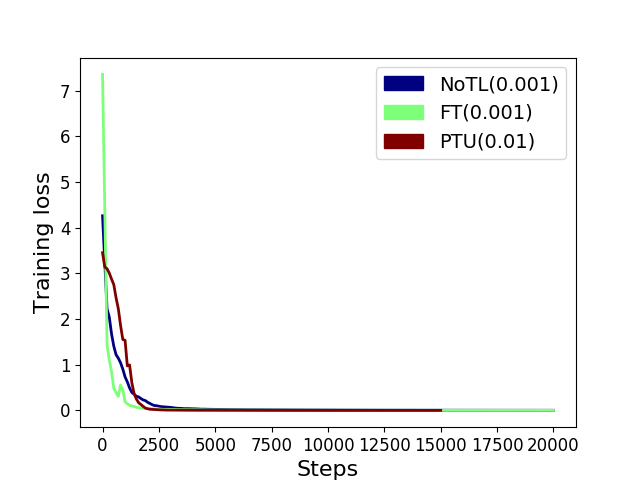}}
  \centerline{\small{(c) Training loss of Greek}}
\end{minipage}
\hfill
\begin{minipage}{.24\linewidth}
  \centerline{\includegraphics[scale=.3,trim={0cm 0cm 0cm 1.2cm},clip]{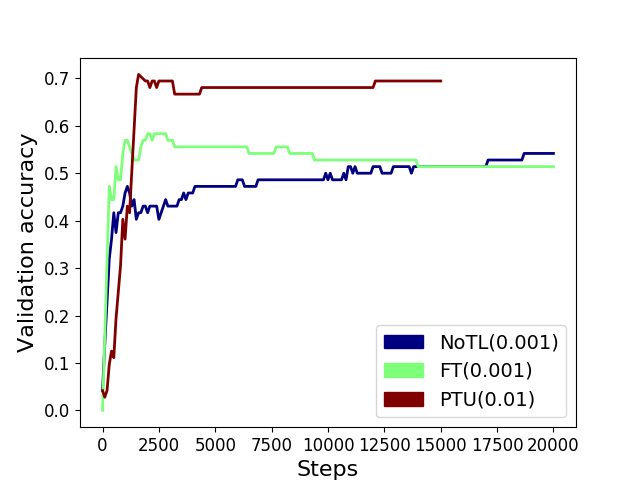}}
  \centerline{\small{(d) Validation accuracy of Greek}}
\end{minipage}
\caption{Learning curves of two transfer settings. }
\label{fig:opt-efficiency}
\end{figure*}

\begin{table*}[htbp]
\centering
\caption{Classification accuracy of MNIST $\rightarrow$ Omniglot}
\label{tab:rnn-cls}
\begin{tabular}{|c|c|c|c|c|c||c|}
\hline
\diagbox[width=2cm]{Datasets}{Models} & RG & K-NN  & NoTL & FT & PTU & $\Delta$ ($\%$) \\ \hline
Greek & $0.0417$ & $0.4028$ & $0.3889$ & $0.4583$ & $\mathbf{0.5139}$ & $12.13$  \\ \hline
Latin & $0.0385$ & $0.3846$ & $0.4103$ & $0.5641$ & $\mathbf{0.6795}$ & $20.46$ \\ \hline
Korean & $0.0250$ & $0.3250$ & $0.3167$ & $0.4833$ & $\mathbf{0.5500}$ & $13.80$  \\ \hline
JP-hiragana & $0.0192$ & $0.4423$ & $0.3590$ & $0.4679$ & $\mathbf{0.5000}$ & $6.86$  \\ \hline
JP-katakana & $0.0213$ & $0.3404$ & $0.2624$ & $0.3404$ & $\mathbf{0.4610}$ & $35.43$ \\
\hline
\end{tabular}
\end{table*}

\subsection{Experiments on RNNs}
The experimental setup is first introduced, and then numerical results are presented. 
\subsubsection{Experimental Setup}
The PTU for RNNs is evaluated with two hand-written character recognition datasets, MNIST \cite{lecun1998gradient} as the source domain and Omniglot \cite{lake2015human} as the target domain. There are $50$ alphabets in the Omniglot dataset, and $5$ alphabets are randomly selected and used as target domains. Each alphabet is composed of a few characters, and there are $20$ labeled samples for each character. The train, validation, test set ratio are $70\%$, $15\%$ and $15\%$ respectively. All the images are resized to $28 \times 28$. At each time step, a row of an image is fed into the RNN. A RNN with $128$ hidden units is used as a classifier. It achieves a classification accuracy at $96\%$ in the source domain. Since the label spaces of the two domains do not agree, the only transferable parameters are those in the RNN cell, and hence there is only one FT strategy. In addition to the NoTL and FT models, a random guess baseline, denoted by RG, and a K-Nearest Neighbor (K-NN) classifier are included as well. The K-NN classifier is implemented with scikit-learn \cite{scikit-learn}. 

RNNs are optimized with the stochastic gradient descent method. The batch size is $128$. The learning rate for training RNNs and the number of neighbors $K$ of the K-NN classifier are treated as hyper-parameters. They are tuned on the validation set. $K$ is selected from $\{1,3,5,10\}$. The learning rate for the NoTL model and the PTU model is selected from $\{0.01, 0.001\}$, and the learning rate for the FT model is selected from $\{0.01, 0.001, 0.0001\}$. 

\subsubsection{Results}
The classification accuracies are listed in Table \ref{tab:rnn-cls}. Since there are only around $1,000$ labeled training data in each target domain, the NoTL model performs even worse than a simple K-NN classifier in $4$ out of $5$ domains. Similar conclusions to the CNN experiments can be drawn. Classification accuracy is improved when a parameter-based transfer learning method is applied, and the proposed PTU further improves over the FT model with a large margin where the relative improvement $\Delta$ ranges from $6.86$ to $35.43$. 

The reason that the PTU outperforms heuristic parameter-sharing and fine-tuning might be two folds. 
\begin{enumerate}
\item The PTU subsumes hand-crafted transfer states by introducing learnable gates. It is more expressive in terms of model capacity. 
\item In PTU, source domain knowledge is retained as frozen parameters, and the domain-specific knowledge is encoded in the target domain network. On the other hand, the parameters in the FT model are changed during training in the target domain, which might impair the useful knowledge from the source domain. 
\end{enumerate}

\subsection{Optimization Efficiency}
We investigate the optimization dynamics of different models by learning curves. The learning curves of two transfer settings, S3 in the CNN experiments and the Greek alphabet as the target domain in the RNN experiments, are shown in Fig. \ref{fig:opt-efficiency}. Both training loss and validation accuracy are reported for each setting, with the learning rate that yields the optimal test classification accuracy. 

For the S3 setting, the NoTL model converges slowly though a large learning rate is used. The validation accuracy is almost $0$ in the first $10,000$ steps. The optimization efficiency is significantly improved by parameter transfer. The FT model converges at a similar rate to the NoTL model while its learning rate is $100$-times smaller. For the PTU model, the training loss drops quickly, and the validation accuracy saturates. The PTU is rather resistant to over-fitting since its validation accuracy does not deteriorate as the training process continues. 

For the Greek setting, the NoTL model gets stuck with a bad local optimum as the training loss decreases while the validation accuracy converges to around $0.5$. The FT model starts with the largest training loss and converges the fastest while it suffers from over-fitting. The PTU model uses a learning rate that is $10$ times larger than the other two baseline models, as it introduces additional parameters. The PTU model has a better generalization ability as it achieves the highest accuracy on the validation set. 

\section{Conclusion}
A principled approach to learn parameter transfer strategy is proposed in this paper. A novel parameter transfer unit (PTU) is designed. The parameter transferability is controlled at a fine-grained scale by two gates in the PTU which can be learned from data. Experimental results demonstrate the effectiveness of the PTU with both CNNs and RNNs in multiple transfer settings where it outperforms heuristic parameter-sharing and fine-tuning. In the future, we will apply the PTU in more challenging settings, for example, image captioning which involves multi-modality data. It is also worth exploring improving existing transfer learning models with the PTU. % gate mechanism quantifies parameter transferability, understanding the behavior of nn. 

\bibliographystyle{aaai}
\bibliography{ijcai18}

\end{document}